\documentclass[letterpaper, 10 pt, conference]{class/ieeeconf}  

\IEEEoverridecommandlockouts                              
\usepackage{amsmath}
\usepackage{amssymb}
\usepackage{latexsym}
\usepackage{url}
\usepackage{cite}
\usepackage{relsize}
\usepackage{multirow}
\usepackage{afterpage}
\usepackage{ifthen}
\usepackage{graphicx}
\usepackage{algpseudocode,algorithm,algorithmicx}
\usepackage{subfigure}
\usepackage[keeplastbox]{flushend}
\usepackage{epstopdf}
\usepackage{stackengine}
\usepackage{gensymb}
\usepackage[bookmarks=false, linkcolor=blue, urlcolor=blue, citecolor=blue]{hyperref} 
\usepackage{booktabs}
\usepackage{makecell}
\usepackage{hhline}
\usepackage{lipsum}
\usepackage{mathtools}

\usepackage[10pt]{moresize}

\hypersetup{pdfstartview={FitH}}

\newcommand \tableCaption[4]{ 
    \begin{tabular}{ll} 
         	\textbf{GT}: #1 	&	\textbf{SGC}: #3\\     
            \textbf{Ours}: #2 	&	\textbf{S2VT}: #4\\    
     \end{tabular}
}

\title{\LARGE \bf Translating Videos to Commands for Robotic Manipulation\\ with Deep Recurrent Neural Networks}

\author{Anh Nguyen$^1$, Dimitrios Kanoulas$^1$, Luca Muratore$^{1,2}$, Darwin G. Caldwell$^1$, and Nikos G. Tsagarakis$^1$
\thanks{$^1$Advanced Robotics Department, Istituto Italiano di Tecnologia, Italy.} 
\thanks{$^2$School of Electrical and Electronic Engineering, The University of Manchester, UK.}
\thanks{{\tt \{Anh.Nguyen, Dimitrios.Kanoulas, Luca.Muratore, Darwin.Caldwell, Nikos.Tsagarakis\}@iit.it}}}

\begin{document}

\newtheorem{problem}{Problem}
\newtheorem{lemma}{Lemma}
\newtheorem{theorem}[lemma]{Theorem}
\newtheorem{claim}{Claim}
\newtheorem{corollary}[lemma]{Corollary}
\newtheorem{definition}[lemma]{Definition}
\newtheorem{proposition}[lemma]{Proposition}
\newtheorem{remark}[lemma]{Remark}
\newenvironment{LabeledProof}[1]{\noindent{\it Proof of #1: }}{\qed}

\def\beq#1\eeq{\begin{equation}#1\end{equation}}
\def\bea#1\eea{\begin{align}#1\end{align}}
\def\beg#1\eeg{\begin{gather}#1\end{gather}}
\def\beqs#1\eeqs{\begin{equation*}#1\end{equation*}}
\def\beas#1\eeas{\begin{align*}#1\end{align*}}
\def\begs#1\eegs{\begin{gather*}#1\end{gather*}}

\newcommand{\poly}{\mathrm{poly}}
\newcommand{\eps}{\epsilon}
\newcommand{\e}{\epsilon}
\newcommand{\polylog}{\mathrm{polylog}}
\newcommand{\rob}[1]{\left( #1 \right)} 
\newcommand{\sqb}[1]{\left[ #1 \right]} 
\newcommand{\cub}[1]{\left\{ #1 \right\} } 
\newcommand{\rb}[1]{\left( #1 \right)} 
\newcommand{\abs}[1]{\left| #1 \right|} 
\newcommand{\zo}{\{0, 1\}}
\newcommand{\zonzo}{\zo^n \to \zo}
\newcommand{\zokzo}{\zo^k \to \zo}
\newcommand{\zot}{\{0,1,2\}}
\newcommand{\en}[1]{\marginpar{\textbf{#1}}}
\newcommand{\efn}[1]{\footnote{\textbf{#1}}}
\newcommand{\vecbm}[1]{\boldmath{#1}} 
\newcommand{\uvec}[1]{\hat{\vec{#1}}}
\newcommand{\thv}{\vecbm{\theta}}
\newcommand{\junk}[1]{}
\newcommand{\var}{\mathop{\mathrm{var}}}
\newcommand{\rank}{\mathop{\mathrm{rank}}}
\newcommand{\diag}{\mathop{\mathrm{diag}}}
\newcommand{\tr}{\mathop{\mathrm{tr}}}
\newcommand{\acos}{\mathop{\mathrm{acos}}}
\newcommand{\atantwo}{\mathop{\mathrm{atan2}}}
\newcommand{\SVD}{\mathop{\mathrm{SVD}}}
\newcommand{\quadf}{\mathop{\mathrm{q}}}
\newcommand{\linterp}{\mathop{\mathrm{l}}}
\newcommand{\sgn}{\mathop{\mathrm{sign}}}
\newcommand{\sym}{\mathop{\mathrm{sym}}}
\newcommand{\avg}{\mathop{\mathrm{avg}}}
\newcommand{\mean}{\mathop{\mathrm{mean}}}
\newcommand{\erf}{\mathop{\mathrm{erf}}}
\newcommand{\grad}{\nabla}
\newcommand{\R}{\mathbb{R}}
\newcommand{\defeq}{\triangleq}
\newcommand{\dims}[2]{[#1\!\times\!#2]}
\newcommand{\sdims}[2]{\mathsmaller{#1\!\times\!#2}}
\newcommand{\udims}[3]{#1}
\newcommand{\udimst}[4]{#1}
\newcommand{\com}[1]{\rhd\text{\emph{#1}}}
\newcommand{\ind}{\hspace{1em}}
\newcommand{\argmin}[1]{\underset{#1}{\operatorname{argmin}}}
\newcommand{\floor}[1]{\left\lfloor{#1}\right\rfloor}
\newcommand{\step}[1]{\vspace{0.5em}\noindent{#1}}
\newcommand{\quat}[1]{\ensuremath{\mathring{\mathbf{#1}}}}
\newcommand{\norm}[1]{\left\lVert#1\right\rVert}
\newcommand{\ignore}[1]{}
\newcommand{\specialcell}[2][c]{\begin{tabular}[#1]{@{}c@{}}#2\end{tabular}}
\newcommand*\Let[2]{\State #1 $\gets$ #2}
\newcommand{\algorithmicbreak}{\textbf{break}}
\newcommand{\Break}{\State \algorithmicbreak}
\newcommand{\ra}[1]{\renewcommand{\arraystretch}{#1}}

\renewcommand{\vec}[1]{\mathbf{#1}} 

\algdef{S}[FOR]{ForEach}[1]{\algorithmicforeach\ #1\ \algorithmicdo}
\algnewcommand\algorithmicforeach{\textbf{for each}}
\algrenewcommand\algorithmicrequire{\textbf{Require:}}
\algrenewcommand\algorithmicensure{\textbf{Ensure:}}
\algnewcommand\algorithmicinput{\textbf{Input:}}
\algnewcommand\INPUT{\item[\algorithmicinput]}
\algnewcommand\algorithmicoutput{\textbf{Output:}}
\algnewcommand\OUTPUT{\item[\algorithmicoutput]}

\maketitle
\thispagestyle{empty}
\pagestyle{empty}

\begin{abstract}
We present a new method to translate videos to commands for robotic manipulation using Deep Recurrent Neural Networks (RNN). Our framework first extracts deep features from the input video frames with a deep Convolutional Neural Networks (CNN). Two RNN layers with an encoder-decoder architecture are then used to encode the visual features and sequentially generate the output words as the command. We demonstrate that the translation accuracy can be improved by allowing a smooth transaction between two RNN layers and using the state-of-the-art feature extractor. The experimental results on our new challenging dataset show that our approach outperforms recent methods by a fair margin. Furthermore, we combine the proposed translation module with the vision and planning system to let a robot perform various manipulation tasks. Finally, we demonstrate the effectiveness of our framework on a full-size humanoid robot WALK-MAN.

\end{abstract}

\section{INTRODUCTION} \label{Sec:Intro}
The ability to perform actions based on observations of human activities is one of the major challenges to increase the capabilities of robotic systems~\cite{Chrystopher09}. Over the past few years, this problem has been of great interest to researchers and remains an active field in robotics~\cite{Ramirez2015}. By understanding human actions, robots may be able to acquire new skills, or perform different tasks, without the need for tedious programming. It is expected that the robots with these abilities will play an increasingly more important role in our society in areas such as assisting or replacing humans in disaster scenarios, taking care of the elderly, or helping people with everyday life tasks.

In this paper, we argue that there are two main capabilities that a robot must develop to be able to replicate human activities: \textit{understanding} human actions, and \textit{imitating} them. The imitation step has been widely investigated in robotics within the framework of learning from demonstration (LfD)~\cite{Brenna2009}. In particular, there are two main approaches in LfD that focus on improving the accuracy of the imitation process: kinesthetic teaching~\cite{Akgun2012} and motion capture~\cite{Koenemann2014}. While the first approach needs the users to physically move the robot through the desired trajectories, the second approach uses a bodysuit or camera system to capture human motions. Although both approaches successfully allow a robot to imitate a human, the number of actions that the robot can learn is quite limited due to the need of using expensively physical systems (i.e., real robot, bodysuit, etc.) to capture the training data~\cite{Akgun2012}~\cite{Koenemann2014}. 

The \textit{understanding} step, on the other hand, receives more attention from the computer vision community. Two popular problems that receive a great deal of interest are video classification~\cite{Karpathy2014} and action recognition~\cite{Simonyan2014}. However, the outputs of these problems are discrete (e.g., the action classes used in~\cite{Simonyan2014} are ``diving", ``biking", ``skiing", etc.), and do not provide further meaningful clues that can be used in robotic applications. Recently, with the rise of deep learning, the video captioning problem~\cite{Venugopalan2016} has become more feasible to tackle. Unlike the classification or detection tasks, the output of the video captioning task is a natural language sentence, which is potentially useful in robotic applications.

\begin{figure}[!t] 
    \centering

     \includegraphics[scale=0.265]{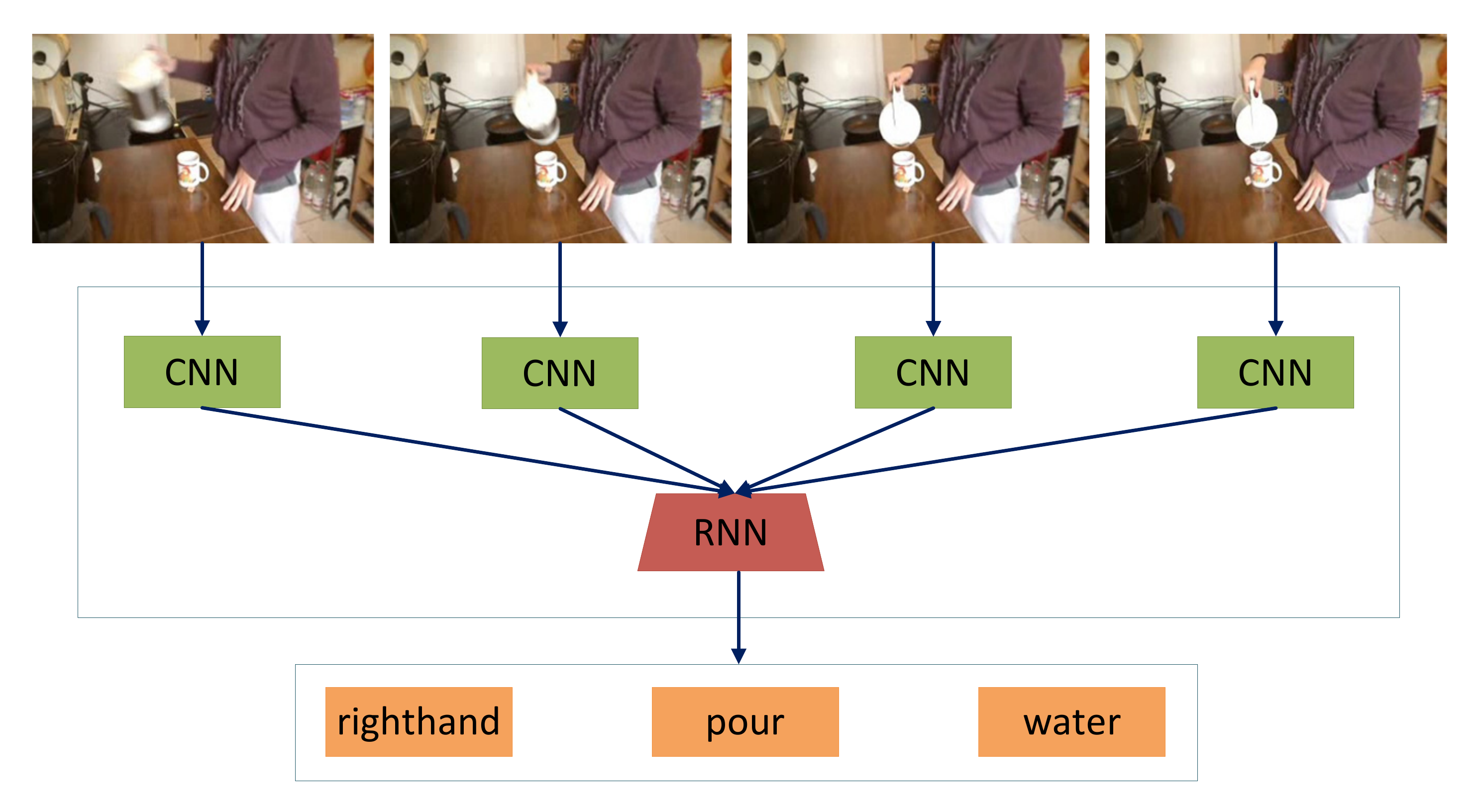}   		
    \vspace{1ex}
    \caption{Translating videos to commands for robotic manipulation. }
    \label{Fig:intro} 
\end{figure}

Inspired by the recent advances in computer vision, this paper describes a deep learning framework that translates an input video to a command that can be used in robotic applications. While the field of LfD~\cite{Brenna2009} focuses mainly on the imitation step, we focus on the understanding step, but our method also allows the robot to perform useful tasks via the output commands. Our goal is to bridge the gap between computer vision and robotics, by developing a system that helps the robot understand human actions, and use this knowledge to complete useful tasks. In particular, we first use CNN to extract deep features from video frames, then two RNN layers are used to learn the relationship between the visual features and the output command. Unlike the video captioning problem~\cite{Venugopalan2016} which describes the output sentence in a natural language form, we use a grammar-free form to describe the output command. We show that our solely neural architecture further improves the state of the art, while its output can be applied in real robotic applications. Fig.~\ref{Fig:intro} illustrates the concept of our method.

Next, we review the related work in Section~\ref{Sec:rw}, then describe our network architecture in Section~\ref{Sec:rnn}. In Section~\ref{Sec:exp}, we present the experimental results on the new challenging dataset, and on the full-size humanoid robot. Finally, we discuss the future work and conclude the paper in Section~\ref{Sec:con}.

\junk{
 actions from studying

The remaining of this paper is organized as follows.
that can be used directly in robotic experiments

directly

the state-of-the-art

discuss the advantages and limitations of our approach then

\begin{itemize}
\item A deep learning based method that can be trained end-to-end to directly interpret input videos to robot commands.

\item A dataset that is sufficient for deep learning methods.

\item An robotic framework that use the proposed method to execute different manipulation tasks.

\end{itemize}

This paper describes the implementation of the ChainModel [15], [16], a biologically inspired model of the mirror neuron system, on an anthropomorphic robot. The work focuses on an interaction and imitation task between the robot and a human, wheremotor control applied to object manipulation is integrated with visual understanding capabilities.More specifically, the re- search focuses on a fully instantiated system integrating perception and learning, capable of executing goal directed motor se- quences and understanding communicative gestures. In particular, the goal is to develop a robotic system that learns to use its motor repertoire to interpret and reproduce observed actions under the guidance of a human user.

A popular approach to solve this problem is learning from demonstration.

The framework of LfD is divided into two fundamental phases: motion transfer and skill modeling. There are two widely used approaches to transfer a motion: kinesthetic guiding and motion capture [4]. The former physically moving the robot in question. This is an intuitive way for users to teach motions to a robot and there are no correspondence issues since the robot is typically aware of its current configuration, i.e., through joint encoders [4,5]. Resulting behaviors tend to be unnatural, because kinesthetic guiding cannot exactly imitate a skill the same way humans would naturally perform it, and it may be difficult for the human to smoothly manipulate the robot. Alternatively, in motion capture systems, users demonstrate a behavior which is extracted by the system, and then converted to a suitable form after some pre-processing steps to remove noise in order to project it onto the robot. One of the main challenges of this approach is how to project recorded data onto the body of the robot, since its morphology may be different from that of the human – the correspondence problem [4]. Typically a body- suit system [6,7] or a camera-based system [8,9] is employed as a motion capturing system. The latter is preferable for demonstrators because it does not constrain performances, allowing for more natural or complex demonstrations.

Transferring skills to humanoid robots based on observations of human activities is widely considered to be one of the most effective ways of increasing the capabilities of such systems [1,2]. It is expected that semantic representations of human activities will play a key role in advancing these sophisticated systems beyond their current capabilities, which will enable these robots to obtain and determine high level understanding of human behaviors. The ability to automatically recognize a human behavior and react to it by generating the next probable motion or action according to human expectations will substantially enrich humanoid robots.

The ability to learn actions from human demonstrations is one of the major challenges for the development of intelligent systems. Particularly, manipulation actions are very challenging, as there is large variation in the way they can be performed and there are many occlusions

This paper describes the implementation of the ChainModel [15], [16], a biologically inspired model of the mirror neuron system, on an anthropomorphic robot. The work focuses on an interaction and imitation task between the robot and a human, wheremotor control applied to object manipulation is integrated with visual understanding capabilities.More specifically, the re- search focuses on a fully instantiated system integrating perception and learning, capable of executing goal directed motor se- quences and understanding communicative gestures. In particular, the goal is to develop a robotic system that learns to use its motor repertoire to interpret and reproduce observed actions under the guidance of a human user.

In order to perform a manipulation action, the robot also needs to learn what tool to grasp and on what object to perform the action. Our system applies CNN based recognition modules to recognize the objects and tools in the video. Then, given the beliefs of the tool and object (from the output of the recognition), our system predicts the most likely action using language, by mining a large corpus using a technique similar to [14]. Putting everything together, the output from the lower level visual perception system is in the form of (LeftHand GraspType1 Object1 Action RightHand GraspType2 Object2). We will refer to this septet of quantities as visual sentence.

At the higher level of representation, we generate a symbolic command sequence. [13] proposed a context-free grammar and related operations to parse manipulation actions. However, their system only processed RGBD data from a controlled lab environment. Furthermore, they did not consider the grasping type in the grammar. This work extends [13] by modeling manipulation actions using a

}

\section{Related Work} \label{Sec:rw}

In the robotic community, LfD techniques are widely used to teach the robots new skills based on human demonstrations. Koenemann et al.~\cite{Koenemann2014} introduced a real-time method to allow a humanoid robot to imitate human whole-body motions. Recently, Welschehold et al.~\cite{Welschehold2016} proposed to transform human demonstrations to different hand-object trajectories in order to adapt to robotic manipulation tasks. The advantage of LfD methods is their abilities to let the robots accurately repeat human motions, however, it is difficult to expand LfD techniques to a large number of tasks since the training process is usually designed for a specific task or needs training data from real robotic systems~\cite{Akgun2012}.

From a computer vision viewpoint, Aksoy et al.~\cite{Aksoy2016} introduced a framework that represents the continuous human actions as ``semantic event chains" and solved the problem as an activity detection task. In~\cite{Yang2015}, Yang et al. proposed to learn manipulation actions from unconstrained videos using CNN and grammar based parser. However, this method needs an explicit representation of both the objects and grasping types to generate command sentences. Recently, the authors in~\cite{Aksoy2017} introduced an unsupervised method to link visual features to textual descriptions in long manipulation tasks. In this paper, we propose to directly learn the output command sentences from the input videos without any prior knowledge. Our method takes advantage of CNN to learn robust features, and RNN to model the sequences, while being easily adapted to any human activity.

Although commands, or in general natural languages, are widely used to control robotic systems. They are usually carefully programmed for each task. This limitation means programming is tedious if there are many tasks. To automatically understand the commands, the authors in~\cite{Tellex2011} formed this problem as a probabilistic graphical model based on the semantic structure of the input command. Similarly, Guadarrama et at.~\cite{Guadarrama13} introduced a semantic parser that used both natural commands and visual concepts to let the robot execute the task. While we retain the concepts of~\cite{Tellex2011} and ~\cite{Guadarrama13}, the main difference in our approach is that we directly use the grammar-free commands from the translation module. This allows us to use a simple similarity measure to map each word in the generated command to the real command on the robot.

In deep learning, Donahue et al.~\cite{Donahue2014} made a first attempt to tackle the video captioning problem. The features were first extracted from video frames with CRF then fed to a LSTM network to produce the output captions. In~\cite{Venugopalan2016}, the authors proposed a sequence-to-sequence model to generate captions for videos from both RGB and optical flow images. Yu et al.~\cite{Haonan2016} used a hierarchical RNN to generate one or multiple sentences to describe a video. In this work, we cast the problem of translating videos to commands as a video captioning task to build on the strong state  the art in computer vision. Furthermore, we use the output of the deep network as the input command to control a full-size humanoid robot, allowing it to perform different manipulation tasks.


\junk{

Recently, the authors in~\cite{Matthias17} introduced a method to solve the problem of linking human whole-body motion and natural language as a captioning task.

, while avoid using the complicated natural language parser
This approach, however, is not straightforward to apply to real robotic applications since the complicated of human spoken language.

This work formed the problem as an activity detection task and did not be applied to robotic applications.

A drawback of this approach ...

The action understanding problem has been extensively studied in robotics and computer vision over the last few years. 

Our work differs in that we are concerned with completing a known task, and focus on when to ask questions as opposed to what type of question to ask

}

\section{Translating Videos to Commands} \label{Sec:rnn}
We start by formulating the problem and briefly describing two popular RNNs use in our method: Long-Short Term Memory (LSTM)~\cite{Hochreiter97_LSTM} and Gated Recurrent Neural network (GRU)~\cite{Cho14_GRU}. Then we present the network architecture that translates the input videos to robotic commands.

\subsection{Problem Formulation}
We cast the problem of translating videos to commands as a video captioning task. In particular, the input video is considered as a list of frames, presented by a sequence of features  $\mathbf{X} = (\mathbf{x}_1, \mathbf{x}_2, ..., \mathbf{x}_n)$ from each frame. The output command is presented as a sequence of word vectors $\mathbf{Y}=(\mathbf{y}_1, \mathbf{y}_2, ..., \mathbf{y}_m)$, in which each vector $\mathbf{y}$ represents one word in the dictionary $D$. The video captioning task is to find for each sequence feature $\mathbf{X}_i$ its most probable command $\mathbf{Y}_i$. In practice, the number of video frames $n$ is usually greater than the number of words $m$. To make the problem become more suitable for robotic applications, we use a dataset that contains mainly human's manipulation actions and assume that the output command $\mathbf{Y}$ is in grammar-free format.

\subsection{Recurrent Neural Networks}

\subsubsection{LSTM}
LSTM is a well-known RNN for effectively modelling the long-term dependencies from the input data. The core of an LSTM network is a memory cell $\mathbf{c}$ which has the gate mechanism to encode the knowledge of the previous inputs at every time step. In particular, the LSTM takes an input $\mathbf{x}_t$ at each time step $t$, and computes the hidden state $\mathbf{h}_t$ and the memory cell state $\mathbf{c}_t$ as follows:

\begin{equation}
\label{Eq_LSTM} 
\begin{aligned} 
{\mathbf{i}_t} &= \sigma ({\mathbf{W}_{xi}}{\mathbf{x}_t} + {\mathbf{W}_{hi}}{\mathbf{h}_{t - 1}} + {\mathbf{b}_i})\\
{\mathbf{f}_t} &= \sigma ({\mathbf{W}_{xf}}{\mathbf{x}_t} + {\mathbf{W}_{hf}}{\mathbf{h}_{t - 1}} + {\mathbf{b}_f})\\
{\mathbf{o}_t} &= \sigma ({\mathbf{W}_{xo}}{\mathbf{x}_t} + {\mathbf{W}_{ho}}{\mathbf{h}_{t - 1}} + {\mathbf{b}_o})\\
{\mathbf{g}_t} &= \phi ({\mathbf{W}_{xg}}{\mathbf{x}_t} + {\mathbf{W}_{hg}}{h_{t - 1}} + {\mathbf{b}_g})\\
{\mathbf{c}_t} &= {\mathbf{f}_t} \odot {\mathbf{c}_{t - 1}} + {\mathbf{i}_t} \odot {\mathbf{g}_t}\\
{\mathbf{h}_t} &= {\mathbf{o}_t} \odot \phi ({\mathbf{c}_t})
\end{aligned}
\end{equation}
where $\odot$ represents element-wise multiplication, the function $\sigma: \mathbb{R} \mapsto [0,1], \sigma (x) = \frac{1}{{1 + {e^{ - x}}}}$ is the sigmod non-linearity, and $\phi: \mathbb{R} \mapsto [ - 1,1], \phi (x) = \frac{{{e^x} - {e^{ - x}}}}{{{e^x} + {e^{ - x}}}}$ is the hyperbolic tangent non-linearity. The weight $W$ and bias $b$ are trained parameters. With this gate mechanism, the LSTM network can remember or forget information for long periods of time, while is still robust against vanishing or exploding gradient problems. In practice, the LSTM network is straightforward to train end-to-end and can handle inputs with different lengths using the padding techniques.

\subsubsection{GRU}
A popular variation of the LSTM network is the GRU proposed by Cho et al.~\cite{Cho14_GRU}. The main advantage of the GRU network is that it requires fewer computations in comparison with the standard LSTM, while the accuracy between these two networks are competitive. Unlike the LSTM network, in a GRU, the update gate controls both the input and forget gates, and the reset gate is applied before the nonlinear transformation as follows:

\begin{figure*}[!htbp] 
    \centering
	\includegraphics[scale=0.31]{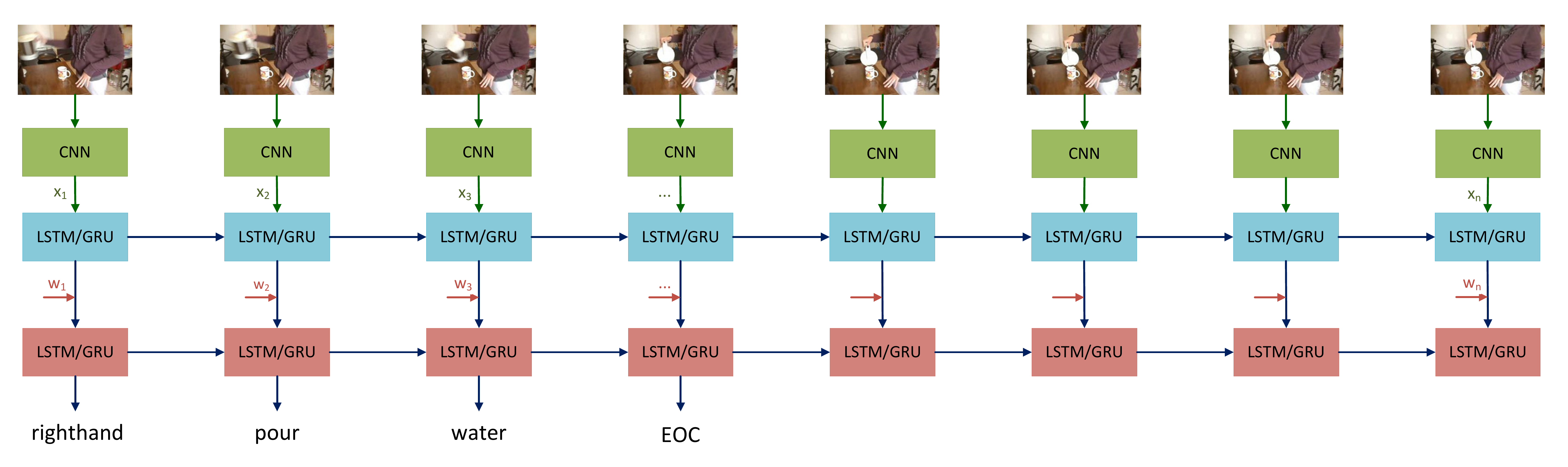} 
    \vspace{1.0ex}
    \caption{An overview of our approach. We first extract the deep features from the input frames using CNN. Then the first LSTM/GRU layer is used to encode the visual features. The input words are fed to the second LSTM/GRU layer and this layer sequentially generates the output words.}
    \label{Fig:overview} 
\end{figure*}

\begin{equation}
\label{Eq_GRU} 
\begin{aligned} 
{\mathbf{r}_t} &= \sigma ({\mathbf{W}_{xr}}{\mathbf{x}_t} + {\mathbf{W}_{hr}}{\mathbf{h}_{t - 1}} + {\mathbf{b}_r})\\
{\mathbf{z}_t} &= \sigma ({\mathbf{W}_{xz}}{\mathbf{x}_t} + {\mathbf{W}_{hz}}{\mathbf{h}_{t - 1}} + {\mathbf{b}_z})\\
{\mathbf{{\tilde h}}_t} &= \phi ({\mathbf{W}_{xh}}{\mathbf{x}_t} + {\mathbf{W}_{hh}}({\mathbf{r}_t} \odot {\mathbf{h}_{t - 1}}) + {\mathbf{b}_h}\\
{\mathbf{h}_t} &= {\mathbf{z}_t} \odot {\mathbf{h}_{t - 1}} + (1 - {\mathbf{z}_t}) \odot {\mathbf{{\tilde h}}_t}
\end{aligned}
\end{equation}
where $\mathbf{r}_t$, $\mathbf{z}_t$, $\mathbf{h}_t$ represent the reset, update, and hidden gate respectively.

\subsection{Videos to Commands}

\subsubsection{Command Embedding} Since a command is a list of words, we have to represent each word as a vector for computation. There are two popular techniques for word representation: \textit{one-hot} encoding and \textit{word2vec}~\cite{Mikolov2013} embedding. Although the one-hot vector is high dimensional and sparse since its dimensionality grows linearly with the number of words in the vocabulary, it is straightforward to use this embedding in the video captioning task. In this work, we choose the one-hot encoding technique as our word representation since the number of words in our dictionary is relatively small (i.e., $|D|=128$). The one-hot vector $\mathbf{y} \in {\mathbb{R}^{|D|}}$ is a binary vector with only one non-zero entry indicating the index of the current word in the vocabulary. Formally, each value in the one-hot vector $\mathbf{y}$ is defined by:

 \begin{equation}
    \mathbf{y}^j=
    \begin{cases}
      1, & \text{if}\ j=ind(\mathbf{y}) \\
      0, & \text{otherwise}
    \end{cases}
  \end{equation}
where $ind(\mathbf{y})$ is the index of the current word in the dictionary $D$. In practice, we add an extra word \textsf{EOC} to the dictionary to denote the end of command sentences.

\subsubsection{Visual Features}
We first sample $n$ frames from each input video in order to extract deep features from the images. The frames are selected uniformly with the same interval if the video is too long. In case the video is too short and there are not enough $n$ frames, we create an artificial frame from the mean pixel values of the ImageNet dataset~\cite{Olga2015} and pad this frame at the end of the list until it reaches $n$ frames. We then use the state-of-the-art CNN to extract deep features from these input frames. Since the visual features provide the key information for the learning process, three popular CNN are used in our experiments: VGG16~\cite{SimonyanZ14}, Inception\_v3~\cite{Szegedy16_Inception}, and ResNet50~\cite{He2016}.

Specifically, for the VGG16 network, the features are extracted from its last fully connected $\texttt{fc2}$ layer. For the Inception\_v3 network, we extract the features from its $\texttt{pool\_3:0}$ tensor. Finally, we use the features from $\texttt{pool5}$ layer of the ResNet50 network. The dimension of the extracted features is $4096$, $2048$, $2048$, for the VGG16, Inception\_v3, and ResNet50 network, respectively. All these CNN are pretrained on ImageNet dataset for image classifications. We notice that the names of the layers we mention here are based on the Tensorflow~\cite{TensorFlow2015} implementation of these networks.

\subsubsection{Architecture}
Our architecture is based on the encoder-decoder scheme~\cite{Venugopalan2016}~\cite{venugopalan2014translating}~\cite{Ramanishka2017cvpr}, which is adapted from the popular sequence to sequence model~\cite{Sutskever2014_Seq} in machine translation. Although recent approaches to video captioning problem use attention mechanism~\cite{Ramanishka2017cvpr} or hierarchical RNN~\cite{Haonan2016}, our proposal solely relies on the neural architecture. Based on the input data characteristics, our network smoothly encodes the input visual features and generates the output commands, achieving a fair improvement over the state of the art without using any additional modules.

In particular, given an input video, we first extract visual features from the video frames using the pretrained CNN network. These features are encoded in the first RNN layer to create the encoder hidden state. The input words are then fed to the second RNN layer, and this layer will decode sequentially to generate a list of words as the output command. Fig.~\ref{Fig:overview} shows an overview of our approach. More formally, given an input sequence of features $\mathbf{X} = (\mathbf{x}_1, \mathbf{x}_2, ..., \mathbf{x}_n)$, we want to estimate the conditional probability for an output command  $\mathbf{Y}=(\mathbf{y}_1, \mathbf{y}_2, ..., \mathbf{y}_m)$ as follows:
\begin{equation} \label{Eq_mainP1}
P(\mathbf{y}_1, ..., \mathbf{y}_m | \mathbf{x}_1, ..., \mathbf{x}_n) = \prod\limits_{i = 1}^m {P({\mathbf{y}_i}|{\mathbf{y}_{i - 1}}, ...,{\mathbf{y}_1}},\mathbf{X})
\end{equation}

Since we want a generative model that encodes a sequence of features and produces a sequence of words in order as a command, the LSTM/GRU is well suitable for this task. Another advantage of LSTM/GRU is that they can model the long-term dependencies in the input features and the output words. In practice, we conduct experiments with the LSTM and GRU network as our RNN, while the input visual features are extracted from the VGG16, Inception\_v3, and ResNet50 network, respectively.

In the encoding stage, the first LSTM/GRU layer converts the visual features $\mathbf{X} = (\mathbf{x}_1, \mathbf{x}_2, ..., \mathbf{x}_n)$ to a list of hidden state vectors $\mathbf{H}^e = (\mathbf{h}_1^e, \mathbf{h}_2^e, ..., \mathbf{h}_n^e)$ (using Equation~\ref{Eq_LSTM} for LSTM or Equation~\ref{Eq_GRU} for GRU). Unlike~\cite{venugopalan2014translating} which takes the average of all $n$ hidden state vectors to create a fixed-length vector, we directly use each hidden vector $\mathbf{h}_i^e$ as the input $\mathbf{x}_i^d$ for the second decoder layer. This allows the smooth transaction from the visual features to the output commands without worrying about the harsh average pooling operation, which can lead to the loss of temporal structure underlying the input video.

In the decoding stage, the second LSTM/GRU layer converts the list of hidden encoder vectors $\mathbf{H}^e$ into the sequence of hidden decoder vectors $\mathbf{H}^d$. The final list of predicted words $\mathbf{\hat{Y}}$ is achieved by applying a softmax layer on the output $\mathbf{H}^d$ of the LSTM/GRU decoder layer. In particular, at each time step $t$, the output $\mathbf{z}_t$ of each LSTM/GRU cell in the decoder layer is passed though a linear prediction layer $\hat{\mathbf{y}}=\mathbf{W}_{z}\mathbf{z}_t+\mathbf{b}_z$, and the predicted distribution $P(\mathbf{y}_t)$ is computed by taking the softmax of $\hat{\mathbf{y}_t}$ as follows:

\begin{equation}
P(\mathbf{y}_t=\mathbf{w}|\mathbf{z}_t)=\frac{\text{exp}(\hat{\mathbf{y}}_{t,w})}{\sum_{w'\in{D}}\text{exp}(\hat{\mathbf{y}}_{t,w'})}
\end{equation}
where $\mathbf{W}_z$ and $\mathbf{b}_z$ are learned parameters, $\mathbf{w}$ is a word in the dictionary $D$.

In this way, the LSTM/GRU decoder layer sequentially generates a conditional probability distribution for each word of the output command given the encoded features representation and all the previously generated words. In practice, we preprocess the data so that the number of input words $m$ is equal to the number of input frames $n$. For the input video, this is done by uniformly sampling $n$ frames in the long video, or padding the extra frame if the video is too short. Since the number of words $m$ in the input commands is always smaller than $n$, we pad a special empty word to the list until we have $n$ words.

\subsubsection{Training}

The network is trained end-to-end with Adam optimizer~\cite{kingma2014adam} using the following objective function:

\begin{equation} \label{Eq_LossFunction}
\mathop {\arg \max }\limits_\theta  \sum\limits_{i = 1}^m {logP({\mathbf{y}_i}|{\mathbf{y}_{i - 1}},...,{\mathbf{y}_1};\theta)} 
\end{equation}
where $\theta$ represents the parameters of the network.

%
%
%

During the training phase, at each time step $t$, the input feature $\mathbf{x}_t$ is fed to an LSTM/GRU cell in the encoder layer along with the previous hidden state $\mathbf{h}_{t-1}^e$ to produce the current hidden state $\mathbf{h}_t^e$. After all the input features are exhausted, the word embedding and the hidden states of the first LSTM/GRU encoder layer are fed to the second LSTM/GRU decoder layer. This decoder layer converts the inputs into a sequence of words by maximizing the log-likelihood of the predicted word (Equation~\ref{Eq_LossFunction}). This decoding process is performed sequentially for each word until the network generates the end-of-command (\textsf{EOC}) token.

\junk{
This layer models the probability distribution of the next word over the word space as follow:
  
The key different in our architecture in comparision with~\cite{venugopalan2014translating} is we do not do average pooling when feeding frame features to LSTM/GRU, instead each feature is fed directly to LSTM/GRU. If the number of frame is less than the number of LSTM step, we pad the missing frames with the mean image from ImageNet dataset. TODO: Compare more.

In the first several time steps, the top LSTM layer (colored red in Figure 2) receives a se- quence of frames and encodes them while the secondLSTM layer receives the hidden representation (ht) and concate- nates it with null padded input words (zeros), which it then encodes. There is no loss during this stage when the LSTMs are encoding. After all the frames in the video clip are ex- hausted, the second LSTM layer is fed the beginning-of- sentence (<BOS>) tag, which prompts it to start decoding its current hidden representation into a sequence of words. While training in the decoding stage, the model maximizes for the log-likelihood of the predicted output sentence given the hidden representation of the visual frame sequence, and the previouswords it has seen. From Equation 3 for a model with parameters θ and output sequence Y = (y1, . . . , ym), this is formulated as

During training the visual feature, sentence pair (V,S) is provided to the model, which then optimizes the log-likelihood (Equation 1) over the entire training dataset using stochastic gradient descent. At each time step, the input xt is fed to the LSTM along with the previous time step’s hidden state ht−1 and the LSTM emits the next hidden state vector ht (and a word). For the first layer of the LSTM xt is the con- catenation of the visual feature vector and the pre- vious encoded word, the ground truth word during training and the predicted word during test time). For the second layer of the LSTM xt is zt of the first layer. Accordingly, inference must also be performed sequentially in the order h1 = fW(x1, 0), h2 = fW(x2,h1), until the model emits the end- of-sentence (EOS) token at the final step T. In our model the output (ht = zt) of the second layer LSTM unit is used to obtain the emitted word

 It updates its hid- den state every time a new word arrives, and encodes the sentence semantics in a compact form up to the words that have been fed in.

 decoder is conditioned step by step on the first t words of the caption and on the corresponding video descriptor, and is tr111ained to produce the next word of the caption. The objective function which we optimize is the log-likelihood of correct words over the sequence

In practice, (THEORY) - PRACTICE These features are then fed to a RNN (LSTM or GRU) to encode them to a fixed length vector $\mathbf{z}$. This process is done by extracting the hidden state vector $h_i^e$ from LSTM/GRU. The middle state is computed by taking the average of all $m$ feature descriptors:

\begin{equation}
z = \frac{1}{m}\sum\limits_{i = 1}^m {{x_i}}
\end{equation}

The decoder converts the encoded vector $z$ into output sequence of words $y_t$, $t \in {1, . . . ,n}$. In particular,

Specifically, for the VGG16 net, the feature map is extracted from its last fully connected layer (i.e. $\textit{fc2} \in {\mathbb{R}^{4096}}$). For the Inception\_v3 network, we extract the features from its $\textit{pool\_3:0} \in {\mathbb{R}^{2048}}$ tensor. Finally, we use the features from $\textit{pool5} \in {\mathbb{R}^{2048}}$ layer of ResNet50.

During the training phase, the frame features and  groundtruth one-hot vectors are used to reproduce these vectors. In the test time, the list of one-hot vectors is generated based on just the input features, which represents the output sentence.

We apply our approach to both still image and video description scenarios, adapting a popular encoder-decoder model for video captioning [22] as our base model.

In this section, we first briefly review two popular RNN that we use in our experiments: Long Short-Term Memory (LSTM)~\cite{Hochreiter97_LSTM} and a variation of the LSTM called Gated Recurrent Units (GRU)~\cite{Cho14_GRU}. We then describe our approach to directly translate videos to robot commands. Fig.~\ref{Fig:overview} shows an overview of our approach.

Commands: list of words (in grammar free format)
Video: list of frames represent a short human action. Although we can describe anything, in this paper, we only focus on manipulation actions of human. More formally, let: 

\begin{equation}
 F = ({f_1},{f_2},...,{f_n})
 \end{equation} 
represent a list of features extracted from the video, and:
\begin{equation}
 C = ({c_1},{c_2},...,{c_m})
 \end{equation} 
 
represent a list of words in a command.

that We first briefly describe the Long-Short Term Memory (LSTM) network, which has shown great success to sequential problems such as machine translation~\cite{s}, image captioning~\cite{•}

Recently, Recurrent Neural Networks (RNNs) have received a lot of attention and achieved impressive results in various applications, including speech recognition [25], image captioning [26] and video description [27]. RNNs have also been applied to facial landmark detection. Very recently, Peng et al. [28] propose a recurrent encoder-decoder network for video-based sequential facial landmark detection. They use recurrent networks to align a sequence of the same face in video frames, by exploiting recurrent learning in spatial and temporal dimensions.

In this section, we propose an approach to translate the input videos to robot commands. Our approach is based on deep RNN models (LSTM and GRU), which have been successfully used in various problems such as machine translation~\cite{Ilya2014}, image captioning~\cite{Donahue2014}, video description~\cite{Venugopalan2016}.

These one-hot vectors are then projected into an embedding space with dimension de by multiplication Weyt with a learned parameter matrix We ∈ Rde×K. The result of a matrix- vector multiplication with a one-hot vector is the column of the matrix corresponding to the index of the single non- zero component of the one-hot vector. We can therefore be thought of as a “lookup table,” mapping each of the K words in the vocabulary to a de-dimensional vector.

plus one additional entry for the <BOS> (beginning of sequence) token which is always taken as y0, the vector $y \int RK$ with a single non-zero component yi = 1 denoting the ith word in the vocabulary. 
a binary vector with exactly one non-zero entry at the position indicating the index of the word in the vocabulary

In addition, the proposed recurrent network in this paper contains multiple Long-Short Term Memory (LSTM) components. For self-containness, we give a brief introduction to LSTM. LSTM is one type of the RNNs, which is attractive because it is explicitly designed to remember information for long periods of time. LSTM takes xt, ht−1 and ct−1 as inputs, ht and ct as outputs

where σ(x) = (1 + e−x)−1 is the sigmoid function. The outputs of the sigmoid functions are in the range of [0, 1], here the smaller values indicate “more probability to forget the previous information” and the larger values indicate “more probability to remember the information”. tanh(x)is the tangent non-linearity function and its outputs are in the range [−1, 1]. [x;h] represents the concatenation of x and h. LSTM has four gates, a forget gate ft, an input gate it, an output gate ot and an input modulation gate gt. ht and ct are the outputs. ht is the hidden unit. ct is the memory cell that can be learned to control whether the previous information should be forgotten or remembered.

At time step t, the input to the bottom-most LSTM is the embedded word from the previous time step yt−1. Input words are encoded as one-hot vectors: vectors $y \int RK$ with a single non-zero component yi = 1 denoting the ith word in the vocabulary, where K is the number of words in the vocabulary, plus one additional entry for the <BOS> (beginning of sequence) token which is always taken as y0, the “previous word” at the first time step (t = 1). These one-hot vectors are then projected into an embedding space with dimension de by multiplication Weyt with a learned parameter matrix We ∈ Rde×K. The result of a matrix- vector multiplication with a one-hot vector is the column of the matrix corresponding to the index of the single non- zero component of the one-hot vector. We can therefore be thought of as a “lookup table,” mapping each of the K words in the vocabulary to a de-dimensional vector.
}

\section{EXPERIMENTS} \label{Sec:exp}
\subsection{Dataset}

Recently, the task of describing video using natural language has gradually received more interest in the computer vision community. Eventually, many video description datasets have been released~\cite{Xu16_MSR_Dataset}. However, these datasets only provide general descriptions of the video and there is no detailed understanding of the action. The captions are also written using natural language sentences which can not be used directly in robotic applications. Motivated by these limitations, we introduce a new \textit{video to command} (IIT-V2C) dataset which focuses on \textit{fine-grained} action understanding~\cite{lea2016learning}. Our goal is to create a new large scale dataset that provides fine-grained understanding of human actions in a grammar-free format. This is more suitable for robotic applications and can be used with deep learning methods.

\textbf{Video annotation} 
Since our main purpose is to develop a framework that can be used by real robots for manipulation tasks, we use only videos that contain human actions. To this end, the raw videos in the Breakfast dataset~\cite{Kuehne14_BF_Dataset} are best suited to our purpose since they were originally designed for activity recognition. We only reuse the raw videos from this dataset and manually segment each video into short clips in a fine granularity level. Each short clip is then annotated with a \textit{command sentence} that describes the current human action.

\textbf{Dataset statistics} 
In particular, we reuse $419$ videos from the Breakfast dataset. The dataset contains $52$ unique participants performing cooking tasks in different kitchens. We segment each video (approximately $2-3$ minutes long) into around $10-50$ short clips (approximately $1-15$ seconds long), resulting in $11,000$ unique short videos. Each short video has a single command sentence that describes human actions. We use $70\%$ of the dataset for training and the remaining $30\%$ for testing. Although our new-form dataset is characterized by its grammar-free property for the convenience in robotic applications, it can easily be adapted to classical video captioning task by adding the full natural sentences as the new groundtruth for each video.

\subsection{Evaluation Metric, Baseline, and Implementation}
\textbf{Evaluation Metric} We report the experimental results using the standard metrics in the captioning task~\cite{Xu16_MSR_Dataset}: BLEU, METEOR, ROUGE-L, and CIDEr. This makes our results directly comparable with the recent state-of-the-art methods in the video captioning field.

\textbf{Baseline} We compare our results with two recent methods in the video captioning field: S2VT~\cite{Venugopalan2016} and SGC~\cite{Ramanishka2017cvpr}. The authors of S2VT used LSTM in the encoder-decoder architecture, while the inputs are from the features of RGB images (extracted by VGG16) and optical flow images (extracted by AlexNet). SGC also used LSTM with encoder-decoder architecture, however, this work integrated a saliency guided method as the attention mechanism, while the features are from Inception\_v3. We use the code provided by the authors of the associated papers for the fair comparison.

\begin{figure*}
\centering
\footnotesize
  \stackunder[2pt]{\includegraphics[width=0.49\linewidth, height=0.09\linewidth]{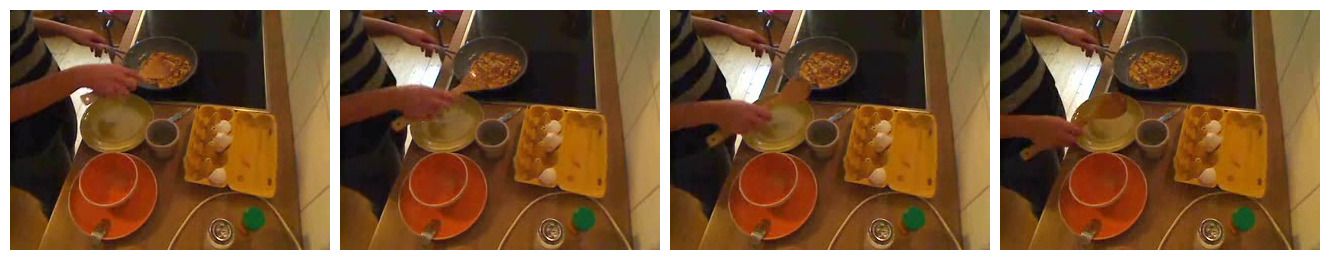}}  				
  				  {\tableCaption {righthand carry spatula} {righthand carry spatula} {lefthand reach stove} {lefthand reach pan}}
  				  \vspace{2ex} 
  \hspace{0.25cm}%
  \stackunder[2pt]{\includegraphics[width=0.49\linewidth, height=0.09\linewidth]{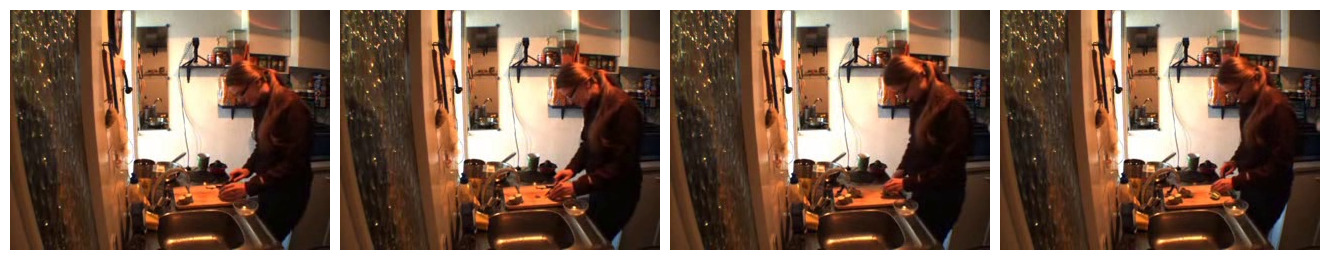}}  
  				  {\tableCaption {righthand cut fruit} {righthand cut fruit} {righthand cut fruit} {righthand cut fruit}}
  				  \vspace{2ex}
  \stackunder[2pt]{\includegraphics[width=0.49\linewidth, height=0.09\linewidth]{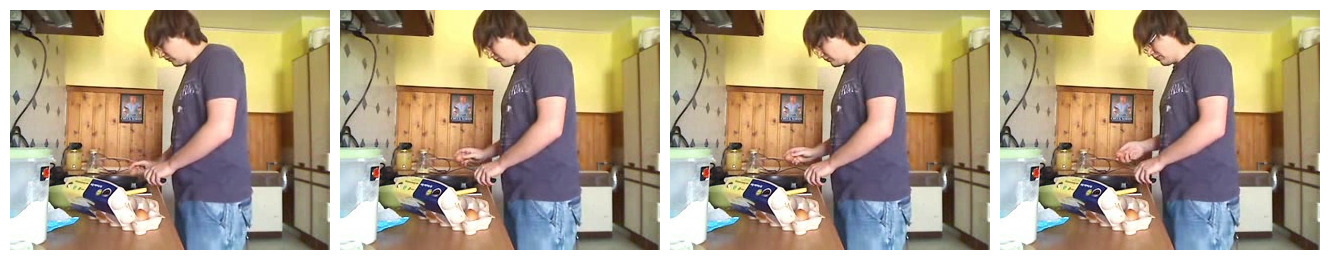}}  
    			  {\tableCaption {righthand crack egg} {righthand carry egg} {lefthand reach spatula} {righthand carry egg}}
  				  \vspace{0ex}
  \hspace{0.25cm}%
  \stackunder[2pt]{\includegraphics[width=0.49\linewidth, height=0.09\linewidth]{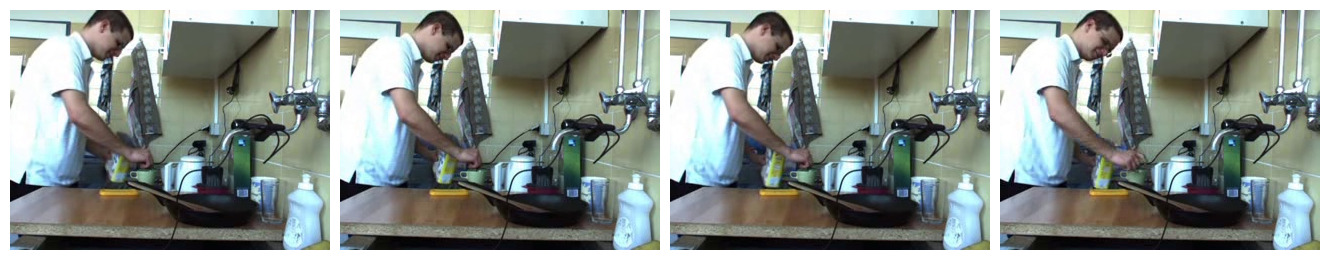}}    
				  {\tableCaption {righthand stir milk} {righthand hold teabag} {righthand place kettle} {righthand take cacao}}
  				  \vspace{0ex}
  
\vspace{1ex}
\caption{Example of translation results of the S2VT, SGC and our LSTM\_Inception\_v3 network on the IIT-V2C dataset.}

\label{Fig:main_result} 
\end{figure*}

\textbf{Implementation} We use $512$ hidden units in both LSTM and GRU in our implementation. The first hidden state of LSTM/GRU is initialized uniformly in $[-0.1, 0.1]$. We set the number of frames for each input video at $30$. Sequentially, we consider each command has maximum $30$ words. If there are not enough $30$ frames/words in the input video/command, we pad the mean frame (from ImageNet dataset)/empty word at the end of the list until it reaches $30$.  During training, we only accumulate the softmax losses of the real words to the total loss, while the losses from the empty words are ignored. We train all the networks for $150$ epochs using Adam optimizer with a learning rate of $0.0001$. The batch size is empirically set to $16$. The training time for each network is around $3$ hours on a NVIDA Titan X GPU.

\subsection{Results}

\begin{table}[!ht]
\centering\ra{1.4}
\caption{Performance on IIT-V2C Dataset}
\renewcommand\tabcolsep{2.5pt}
\label{tb_result_v2c}
\hspace{2ex}

\begin{tabular}{@{}rcccccccc@{}}
\toprule 					 &
\ssmall Bleu\_1  & 
\ssmall Bleu\_2  & 
\ssmall Bleu\_3  &
\ssmall Bleu\_4  & 
\ssmall METEOR  &
\ssmall ROUGE\_L &
\ssmall CIDEr \\

\midrule
S2VT~\cite{Venugopalan2016} 				& 0.383   & 0.265   & 0.201	& 0.159	& 0.183    & 0.382   & 1.431     \\
SGC~\cite{Ramanishka2017cvpr}			& 0.370   & 0.256   & 0.198	& 0.161	& 0.179    & 0.371   & 1.422     \\
\cline{1-8}
LSTM\_VGG16					& 0.372   & 0.255  			 & 0.193	& 0.159	& 0.180    & 0.375   & 1.395     \\
GRU\_VGG16 					& 0.350   & 0.233 			 & 0.173	& 0.137	& 0.168    & 0.351   & 1.255     \\
LSTM\_Inception\_v3				& \textbf{0.400}  			 & \textbf{0.286}   & 0.221	& 0.178	& \textbf{0.194}    & \textbf{0.402}   & \textbf{1.594}     \\
GRU\_Inception\_v3 				& 0.391   & 0.281  			 & \textbf{0.222}	& \textbf{0.188}	& 0.190    & 0.398   & 1.588     \\
LSTM\_ResNet50 				& 0.398   & 0.279            & 0.215	& 0.174	& 0.193    & 0.398   & 1.550     \\
GRU\_ResNet50 				& 0.398   & 0.284   & 0.220	& 0.183	& 0.193    & 0.399   & 1.567     \\
\bottomrule
\end{tabular}
\end{table}

Table~\ref{tb_result_v2c} summarizes the captioning results on the IIT-V2C dataset. Overall, the LSTM network that uses visual features from Inception\_v3 (LSTM\_Inception\_v3) achieves the highest performance, winning on the Blue\_1, Blue\_2, METEOR, ROUGE\_L, and CIDEr metrics. Our LSTM\_Inception\_v3 also outperforms S2VT and SGC in all metrics by a fair margin. We also notice that both the LSTM\_ResNet50 and GRU\_ResNet50 networks give competitive results in comparison with the LSTM\_Inception\_v3 network. Overall, we observe that the architectures that use LSTM give slightly better results than those using GRU. However, this difference is not significant when the ResNet50 features are used to train the models (LSTM\_ResNet50 and GRU\_ResNet50 results are a tie).

From the experiments, we notice that there are two main factors that affect the results of this problem: the network architecture and the input visual features. Since the IIT-V2C dataset contains mainly the fine-grained human actions in a limited environment (i.e., the kitchen), the SGC architecture that used saliency guide as the attention mechanism does not perform well as in the normal video captioning task. On the other hand, the visual features strongly affect the final results. Our experiments show that the ResNet50 and Inception\_v3 features significantly outperform the VGG16 features in both LSTM and GRU networks. Since the visual features are not re-trained in the sequence to sequence model, in practice it is crucial to choose the state-of-the-art CNN as the feature extractor for the best performance.

Fig.~\ref{Fig:main_result} shows some examples of the generated commands by our LSTM\_Inception\_v3, S2VT, and SGC models on the test videos of the IIT-V2C dataset. These qualitative results show that our LSTM\_Inception\_v3 gives good predictions in many cases, while S2VT and SGC results are more variable. In addition to the good predictions that are identical with the groundtruth, we note that many other generated commands are relevant. Due to the nature of the IIT-V2C dataset, most of the videos are short and contain fine-grained human manipulation actions, while the groundtruth commands are also very short. This makes the problem of translating videos to commands is more challenging than the normal video captioning task since the network has to rely on the minimal information to predict the output command.

\subsection{Robotic Applications}

\begin{figure*}[ht]
  \centering
 \subfigure[Pick and place task]{\label{fig_resize_map_a}\includegraphics[width=0.99\linewidth, height=0.16\linewidth]{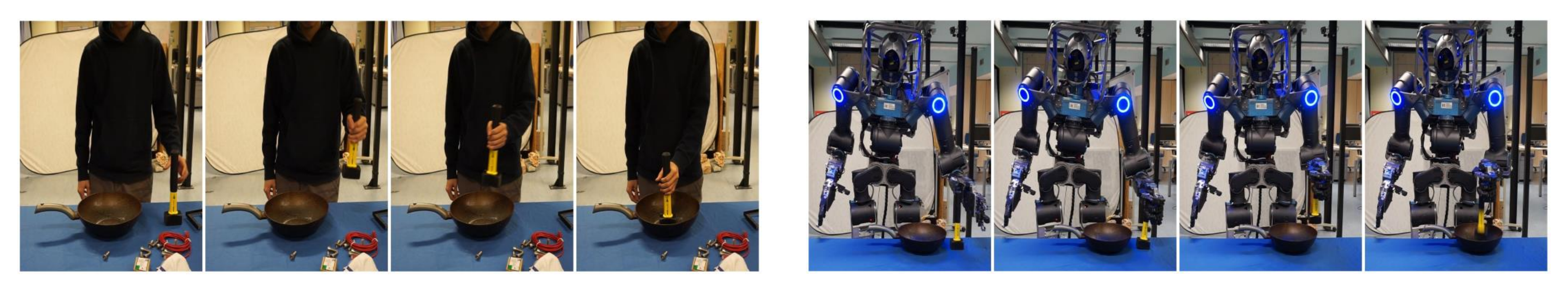}}
 \subfigure[Pouring task]{\label{fig_resize_map_b}\includegraphics[width=0.99\linewidth, height=0.16\linewidth]{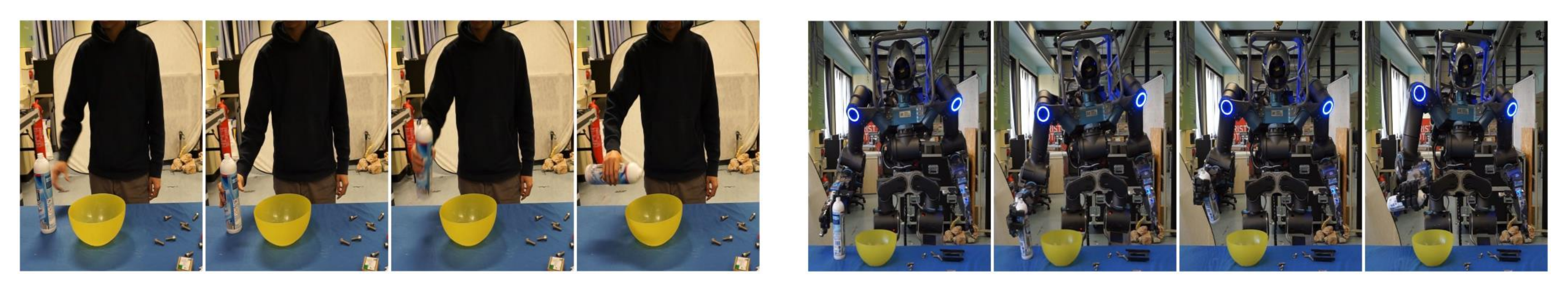}}

 \vspace{2.0ex}
 \caption{Example of manipulation tasks performed by WALK-MAN using our proposed framework. \textbf{(a)} Pick and place task. \textbf{(b)} Pouring task. The frames from human instruction videos are on the left side, while the robot performs actions on the right side. We notice that there are two sub-tasks (i.e., two commands) in these tasks: grasping the object and manipulating it. More illustrations can be found in the supplemental video.}
 \label{Fig:robot_imitation}
\end{figure*}

Given the proposed translation module, we build a robotic framework that allows the robot to perform various manipulation tasks by just ``\textit{watching}" the input video. Our goal in this work is similar to~\cite{Yang2015}, however, we propose to keep the video understanding separately from the vision system. In this way, the robot can learn to understand the task and execute it independently. This makes the proposed approach more practical since it does not require a dataset that has both the caption and the object (or grasping) location. It is also important to note that our goals differ from LfD since we only focus on finding a general way to let the robot execute different manipulation actions, while the trajectory in each action is assumed to be known.

In particular, for each task presented by a video, the translation module will generate an output command sentence. Based on this command, the robot uses its vision system to find relevant objects and plan the actions. Experiments are conducted using the humanoid WALK-MAN~\cite{Niko2016_full}. The robot is controlled using the XBotCore software architecture~\cite{muratore2017xbotcore}, while the OpenSoT library~\cite{Rocchi15} is used to plan full-body motion. The relevant objects and their affordances are detected using AffordanceNet framework~\cite{AffordanceNet17}. For simplicity, we only use objects in the IIT-Aff dataset~\cite{Nguyen2017_Aff} in the demonstration videos so the robot can recognize them. Using this setup, the robot can successfully perform various manipulation tasks by closing the loop: understanding the human demonstration from the video using the proposed method, finding the relevant objects and grasping poses~\cite{Nguyen2017_Aff}, and planning for each action~\cite{Rocchi15}.

Fig.~\ref{Fig:robot_imitation} shows some manipulation tasks performed by WALK-MAN using our proposed framework. For a simple task such as ``righthand grasp bottle", the robot can effectively repeat the human action through the command. Since the output of our translation module is in grammar-free format, we can directly map each word in the command sentence to the real robot command. In this way, we avoid using other modules as in~\cite{Tellex2011} to parse the natural command into the one that uses in the real robot. The visual system also plays an important role in our framework since it provides the object location and the target frames (e.g., grasping frame, ending frame) for the robot to plan the actions. Using our approach, the robot can also complete long manipulation tasks by stacking a list of demonstration videos in order for the translation module. Note that, for the long manipulation tasks, we assume that the ending state of one task will be the starting state of the next task. Overall, WALK-MAN successfully performs various manipulation tasks such as grasping, pick and place, or pouring. The experimental video and our IIT-V2C dataset can be found at the following link:
\vspace{1ex}
\centerline{\url{https://sites.google.com/site/video2command/}}

\junk{
 Furthermore, we can also  We can also stack several videos to create a chain of commands. 
 
Using our approach, the robot can perform different tasks based on the instructions from human demonstration videos.

  can understand the human instructions from the video via the proposed translation module, while the visual information can be solved effectively with the recent advances in deep learning~\cite{Nguyen2017_Aff}.

, and we already have many separated datasets for video captioning~\cite{Xu16_MSR_Dataset} and affordance detection~\cite{Nguyen2017_Aff}

 using all standard metrics
 
 We train each LSTM/GRU network using the features from VGG16, Inception\_v3 and ResNet50, respectively.
 
 method for video captioning and the recent method that used encoder-decoder scheme with saliency guided~\cite{Ramanishka2017cvpr} (denoted as SGC) as the attention mechanism.

   \textbf{Grasping}, \textbf{Pick and Place} In this experiment, we address how the robot executes a new scenario, i.e. ``pick up an object and place it into another object". Our experiment involves two affordances \texttt{grasp} and \texttt{contain}. Using our approach, the robot can grasp an object from its \texttt{grasp} affordance and bring it to a new location that belongs to the \texttt{contain} affordance of another object. Furthermore, we note that with the aid of our semantic understanding framework, the robot can recognize the target objects while ignoring the irrelevant ones. \textbf{Pouring} Similar to the pick and place experiment, we use two affordances \texttt{grasp} and \texttt{contain} and the associated objects in our dataset. The goal is to pour the liquid from an object to the \texttt{contain} region of another object. Our semantic perception framework gives the robot detail understanding about the target objects, their affordances as well as the relative coordinates for the movement. We notice that besides the basic actions such as grasp, raise, etc., we predefined the pouring action to help the robot complete the task. 
   
 we use the vision system from~\cite{Nguyen2017_Aff} to provide object location, its affordances and grasping frame for the robot, while trajectory is generated by the OpenSoT library~\cite{Rocchi15}. For simplicity, we only use the objects that the IIT-Aff~\cite{Nguyen2017_Aff} dataset so the robot can recognize them. Since there are a huge gap between the number of words in the IIT-V2C dataset  and the the number of objects that the robot can recoginze in the  in the  . videoFor each To create the command for  n our work, instead of using full natural command sentences, we propose to use human demonstrations from videos as the input. The translation module is then used to interpret the video to commands in grammar-free format that the robot can follow. Combined with the vision and planning system, our approach allows the robot to perform manipulation tasks by just ``watching" the input video. Fig.~\ref{Fig:robot_apps} shows a full description of framework in our robotic applications.

We validate our framework using the WALK-MAN full-size humanoid~\cite{Niko2016_full}. The robot is controlled in real-time using the XBotCore software architecture~\cite{muratore2017xbotcore}, while the OpenSoT library~\cite{Rocchi15} is used to plan full body motion. The relevant objects and their affordances are detected using the framework in~\cite{Nguyen2017_Aff}. To allow the real-time performance, the control and planning system run a control pc, while the vision system runs on a vision pc with a NVIDIA Titan X GPU.

For the safety of the robot, we define some key action such as "reach", "grasp". Each word in the output command will be compared with these basic action using the similarity in word2vector. 

Fig.~\ref{Fig:robot_apps} shows an overview of our robotic application.

To compare the origi- nality in generation, we compute the Levenshtein distance of the predicted sentences with those in the training set. From Table 3, for the MSVD corpus, 42.9 of the predic- tions are identical to some training sentence, and another 38.3 can be obtained by inserting, deleting or substituting one word from some sentence in the training corpus. We note that many of the descriptions generated are relevant. TODO: challenging, failures case.

 We follow the standard procedure in the captioning tasks to evaluate our results.

 In particular, we use the code from COCO evaluation server~\cite{Chen_COCO_Evaluation} that implements several metrics:

 the baseline network architecture remains challenging to modify, most of the recent work used different models such as attention mechanism~\cite{•}, saliency

GoogLeNet [32, 12] to extract the frame-level features in our experiment. All the videos’ lengths are kept to 200 frames. For a video with more than 200 frames, we drop the extra frames. For a video with- out enough frames, we pad zero frames. These are com- mon approaches to ensure all the videos have the same length [38, 43]. Feature Extractor: CNN. We consider the input video as a sequence of frames and encode each frame using a CNN. This process extracts the meaningful features from the input images at every time step. These features are then fed to the LSTM network as inputs. In particular, we use three most popular CNNs: VGG16, GoogleNet, and ResNet-50 as our feature extractors. For the VGG16 and ResNet-50, we remove the last classification layer and TODO: describe net without the last layer.
\\

The MSR-VTT dataset is characterized by the unique properties including the large scale clip-sentence pairs, comprehensive video categories, diverse video content and descriptions, as well as multimodal audio and video stream- s. We

Current datasets for video to text mostly focus on specific fine-grained domains. For example, YouCook [5], TACoS [25, 28] and TACoS Multi-level [26] are mainly de- signed for cooking behavior. MSR-VTT focuses on general videos in our life, while MPII-MD [27] and M-VAD [32] on movie domain. Although MSVD [3] contains general web videos which may cover different categories, the very limited size (1,970) is far from representativeness. To col- lect representative videos, we obtain the top 257 represen- tative queries from a commercial video search engine, cor- responding to 20 categories

Since our goal is to collect short video clips that each can be described with one single sentence in our current version

. TOFIX: Similar to the treatment of frame features, we embed words to a lower 500 dimensional space by applying a linear transformation to the input data and learning its parameters via back propa- gation. The embedded word vector concatenated with the output (ht) of the first LSTM layer forms the input to the second LSTM layer (marked green in Figure 2). When considering the output of the LSTM we apply a softmax over the complete vocabulary as in Equation 5.

}
\section{Conclusions and Future Work}\label{Sec:con}
In this paper, we proposed a new method to translate human demonstration videos to commands using deep recurrent neural networks. We conducted experiments with the LSTM and GRU network using different visual feature representations. The experimental results showed that our purely neural sequence to sequence architecture outperformed current state-of-the-art methods by a fair margin. We also introduced a new large-scale videos to commands dataset that is suitable for deep learning methods. Finally, we combined our proposed method with the vision and planning module, and performed various manipulation tests on a real full-size humanoid robot.

Our robotic experiments so far are qualitative. We have focused on demonstrating how our approach can be used in a real robotic system to reduce the (tedious) programming when there are many manipulation tasks. Although using the learning approach to translate the demonstration videos to commands could help the robot understand human actions in a meaningful way, the imitation step is still challenging since it requires a robust vision, planning (and LfD) system. Currently, our framework relies solely on the vision system to plan the actions. This does not allow the robot to perform accurate tasks such as ``hammering" or ``cutting" which require precise skills. Therefore, an interesting problem is to combine our approach with LfD techniques to improve the robot manipulation capabilities.

\section*{Acknowledgment}
\addcontentsline{toc}{section}{Acknowledgment}
This work is supported by the European Union Seventh Framework Programme [FP7-ICT-2013-10] under grant agreement no 611832 (WALK-MAN). 

\junk
{

 . task such as  We can also combine our approach with LfD techniques to allow the robot to perform more   It is clear that we need to improve both three components in our framework (translation, vision, and planning) in order to

achieve the human level in manipulation tasks.

the robot can understand the concept of human action using our translation module, however, the imitation step is very challenging since it requires a perfect vision and planning to achieve human level.

We propose the idea of translating videos to commands and use these commands in robotic manipulation tasks. Our approach potentially can perform useful tasks without the need of tedious programming. We have demonstrated that our framework can be used together with other modules such as vision recognition system to complete complicated task. It is also straightforward to combine our framework with other LfD methods to perform tasks that require precise accuracy.

 Using the proposed method, we introduced  detect object affordances with CNN. We have demonstrated that the affordance detection results can be improved by using an object detector and dense CRF. Moreover, we introduced a challenging dataset that is suitable for real-world robotic applications. From the detected affordances, we presented a grasping method that is robust to noisy data. The effectiveness of our approach was demonstrated by performing different grasping experiments in cluttered scenes on the full-size humanoid robot WALK-MAN. We hope this opens up the door to practical solutions to the current limitations of real-world SLAM applications. It is worth restating that t...

Currently, our approach needs two separate networks to detect object affordances. This architecture can't be trained end-to-end as a single network. In future work, we aim to overcome this limitation by developing a new architecture that can detect the object identity and its affordances simultaneously. Another interesting problem is to extend our robotics experiments with more complicated scenarios.
}

\bibliographystyle{class/IEEEtran}
\bibliography{class/IEEEabrv,class/reference}
   
\end{document}